\documentclass[letterpaper, 10 pt, conference]{ieeeconf}  

\IEEEoverridecommandlockouts                              

\usepackage{amsmath,amsfonts}
\usepackage[noend]{algorithmic}
\usepackage{algorithm}
\usepackage{array}
\usepackage[caption=false,font=normalsize,labelfont=sf,textfont=sf]{subfig}
\usepackage{textcomp}
\usepackage{stfloats}
\usepackage{url}
\usepackage{verbatim}
\usepackage{graphicx}
\usepackage{cite}
\usepackage{xcolor}
\usepackage{hyperref}
\usepackage{multirow}
\usepackage{booktabs}
\usepackage{pdfpages}
\usepackage{graphicx}
\usepackage{dsfont}
\usepackage{pifont}

\definecolor{mydarkblue}{rgb}{0,0.08,0.45}
\hypersetup{
    colorlinks=true,
    linkcolor=mydarkblue,
    citecolor=mydarkblue,
    filecolor=mydarkblue,
    urlcolor=mydarkblue,
    }

\begin{document}

\title{\LARGE \bf QuietPaw: Learning Quadrupedal Locomotion with \\Versatile Noise Preference Alignment
}

\author{Yuyou Zhang$^{1}$, 
Yihang Yao$^{1}$, 
Shiqi Liu$^{1}$, 
Yaru Niu$^{1}$,
Changyi Lin$^{1}$, \\
Yuxiang Yang$^{2}$,
Wenhao Yu$^{2}$,
Tingnan Zhang$^{2}$,
Jie Tan$^{2}$,
Ding Zhao$^{1}$
\thanks{$^{1}$Carnegie Mellon University, $^{2}$Google DeepMind. Correspondance to Yuyou Zhang {\tt\small yuyouz@andrew.cmu.edu}.} 
}%

\maketitle

\begin{abstract}
When operating at their full capacity, quadrupedal robots can produce loud footstep noise, which can be disruptive in human-centered environments like homes, offices, and hospitals.
As a result, balancing locomotion performance with noise constraints is crucial for the successful real-world deployment of quadrupedal robots. 
However, achieving adaptive noise control is challenging due to (a) the trade-off between agility and noise minimization, (b) the need for generalization across diverse deployment conditions, and (c) the difficulty of effectively adjusting policies based on noise requirements. 
We propose QuietPaw, a framework incorporating our Conditional Noise-Constrained Policy (CNCP), a constrained learning-based algorithm that enables flexible, noise-aware locomotion by conditioning policy behavior on noise-reduction levels.
We leverage value representation decomposition in the critics, disentangling state representations from condition-dependent representations and this allows a single versatile policy to generalize across noise levels without retraining while improving the Pareto trade-off between agility and noise reduction. 
We validate our approach in simulation and the real world, demonstrating that CNCP can effectively balance locomotion performance and noise constraints, achieving continuously adjustable noise reduction.
\end{abstract}


\section{Introduction}

Recent progress in reinforcement learning (RL)~\cite{lee2020learning,su2024leveraging, margolis2023walk,kim2024not,rudin2022learning} has enabled quadrupedal robots to navigate real-world environments with greater agility and adaptability~\cite{kumar2021rma, hoeller2024anymal, chane-sane2024soloparkour, chen2024slr, long2024learning, mitchell2024gaitor, ren2024topnav, yang2024generalized, lin2024locoman}.
One critical but often overlooked aspect of real-world deployment is noise generation with foot contacts. 
In environments such as hospitals, offices, or residential areas, excessive locomotion noise can be disruptive and undesirable.
However, existing quadrupedal locomotion policies primarily focus on optimizing agility~\cite{he2024agile, yang2023cajun, yang2024agile}, stability~\cite{long2024learning}, and energy efficiency ~\cite{liang2024adaptive}, without explicitly accounting for noise constraints. 

\begin{figure}[t]
    \centering
    \includegraphics[width=1.0\linewidth]{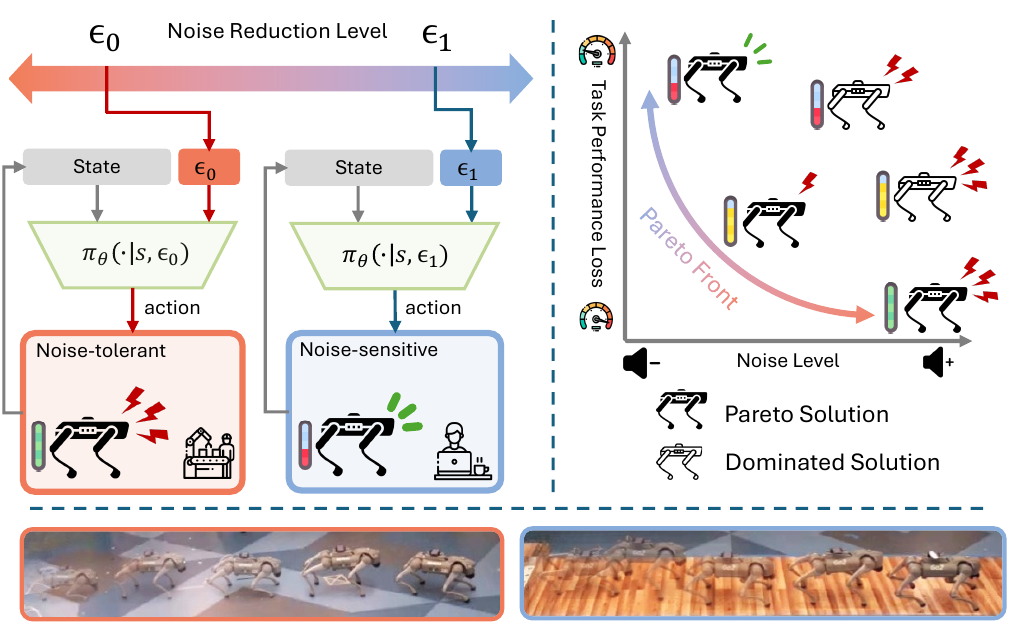}
    \caption{Overview of our Conditional Noise-Constrained Policy (CNCP).  
    \textbf{Left:} CNCP conditions its policy on a noise reduction level $\epsilon$, dynamically adjusting locomotion behavior for different noise sensitivity requirements.  
    \textbf{Right:} The trade-off between task performance and noise reduction forms a Pareto front, where improving one objective compromises the other.  
    Solutions on the Pareto front achieve optimal trade-offs, while dominated solutions are suboptimal in both noise minimization and performance.
    \textbf{Bottom:} The same policy produces different noise levels depending on terrain type, highlighting the necessity for adaptable noise control.}
    \label{fig:overview}
    \vspace{-10pt}
\end{figure}

Minimizing the quadrupedal robot locomotion sound presents several challenges. 
The first challenge arises from the inherent trade-off between optimal locomotion performance and noise minimization. 
Reducing impact forces often leads to less aggressive movements, which in turn can limit agility, stability, or speed.
A policy optimized for task performance may prioritize fast, accurate gaits while neglecting noise constraints, while a purely noise-constrained policy may result in slow, inefficient movement. 
To balance these competing objectives, we employ Constrained RL, modeling noise as a constraint while optimizing locomotion performance within the specified noise limit. 
This approach ensures the policy maximizes locomotion task performance while achieving the desired noise reduction, as illustrated in Figure~\ref{fig:overview}.

However, enforcing a fixed single noise constraint is insufficient.
For a noise-constrained quadrupedal policy to be practically deployable, it must also generalize efficiently to different noise preferences without requiring retraining. 
The second challenge is developing a versatile policy that can be flexibly adjusted to different noise reduction requirements. 
The same action sequence can result in vastly different noise levels depending on the terrain (e.g., soft carpet vs. hard pavement), and different users or applications may have varying noise tolerance thresholds. This variability makes it impractical to design a single fixed policy that meets all possible noise reduction demands. 

Most constrained RL policies assume a fixed single constraint and lack such flexibility to accommodate new constraints. 
However, retraining the policy for every new noise tolerance level or deployment scenario will be highly data-inefficient and impractical.
To overcome this limitation, we adopt a versatile RL locomotion policy that integrates the noise reduction level as a conditioning variable within a constrained RL framework. 
This allows the policy to dynamically adjust locomotion behavior based on the desired noise constraint, enabling real-time adaptability without requiring retraining, as depicted in Figure~\ref{fig:overview}.

Even with a conditional policy, effectively steering locomotion behavior based on noise constraints remains challenging.   
Simply concatenating the noise constraint into the policy input, often fails to induce meaningful behavioral differences~\cite{yang2019generalized, ma2020universal}, as the policy struggles to learn the nuanced trade-offs required for both agility and noise reduction. 
To address this, we incorporate decomposed representation learning into RL for noise-constrained quadrupedal locomotion. By leveraging successor features (SFs)~\cite{nemecek2021policy, alegre2022optimistic} in the value representation, our method separates condition-invariant state features from constraint-dependent adaptations, enabling seamless generalization across different noise levels. 
By doing so, the policy achieves a superior Pareto-optimal trade-off~\cite{deb2011multi, van2014multi} between locomotion performance and noise minimization, ensuring that varying noise constraints result in distinct and effective noise reduction behaviors.

In this work, we introduce \textbf{CNCP}, a RL-based locomotion policy that dynamically adapts to different noise constraints using a successor features decomposed representation.
Our key contributions include:
\begin{itemize}
    \item A conditional RL framework for noise-constrained locomotion that enables adjusting noise reduction without retraining, enhancing adaptability for real-world deployment.
    \item A decomposed value estimation approach that improves Pareto-optimality and ensures effective policy steering based on noise constraints.
    \item Extensive simulation and real-world experiments demonstrating CNCP’s ability to flexibly minimize noise while maintaining locomotion performance, advancing the Pareto front across diverse environments and constraints.
\end{itemize}

\section{Related work}

\subsection{Acoustic-aware Robot}
Nonverbal sound has been explored in human-robot interaction for localization, communication, and sociability \cite{zhang2023nonverbal}. Prior work has also investigated noise-aware path planning for drones, demonstrating the benefits of integrating acoustic constraints into motion planning to reduce human noise exposure \cite{adlakha2023integration}. Additionally, research on energy-efficient and low-impact robotic locomotion highlights the importance of minimizing foot-ground collisions to achieve quieter operation \cite{gomes2015quiet}.  
Most relevant to our work, learning-based approaches have been proposed to reduce noise in robotic locomotion. Watanabe et al. \cite{watanabe2025learning} developed an RL framework for minimizing foot contact velocity in quadrupeds, achieving quieter walking through sim-to-real training. 
However, their method optimizes for a fixed noise objective, limiting adaptability. Inspired by multi-objective reinforcement learning for personalized robotic behaviors \cite{Hwang_2024_CVPR}, we propose an adjustable noise reduction framework that dynamically adapts noise constraints based on deployment conditions.

\subsection{Constrained policy learning} 
Constrained policy learning, also known as safe reinforcement learning, is formulated as a constrained optimization problem that aims to maximize reward performance while satisfying predefined constraints~\cite{achiam2017constrained}. This problem can be solved using Lagrangian-based approaches~\cite{bhatnagar2012online, chow2017risk, as2022constrained, ding2023provably}. Some studies approximate the constrained RL problem using Taylor expansions~\cite{achiam2017constrained} or variational inference~\cite{liu2022constrained}. Another line of research frames it as a multi-objective optimization (MOO) problem~\cite{liu2023constrained} or a multi-task learning problem~\cite{yao2024gradient}. Most of these approaches assume a fixed, predefined safety level during both training and deployment. 
Recently, some works have explored dynamic safety constraints during deployment~\cite{yao2023constraint, chemingui2024constraint}, but those methods focus on reletive simple problems in simulation. Real world robotic applications remain underexplored.

\subsection{RL-Based Quadrupedal Locomotion}
RL has enabled quadrupedal robots to achieve remarkable locomotion capabilities across diverse environments, including robust walking over challenging terrains~\cite{choi2023learning, lee2020learning, kumar2021rma, lindqvist2022multimodality, miki2024learning}, jumping~\cite{yang2023cajun, yang2024agile}, dribbling~\cite{ji2023dribblebot}, pushing~\cite{nachum2019multi, feng2024learning}, climbing~\cite{vogel2024robust}, parkour~\cite{zhuang2023robot, cheng2024extreme, hoeller2024anymal}, and high-speed running~\cite{yang2022fast, he2024agile}. While most existing approaches emphasize robustness, agility, dexterity, energy efficiency, or collision avoidance, they often overlook the user-friendliness of locomotion policies. In this work, we propose a constrained RL approach to regulate the noise level of the learned policy while minimizing performance degradation, enabling user-friendly locomotion across various scenarios.

\section{Method}

\subsection{Noise-Constrained Locomotion with Safe RL}
\label{method: formulation}

To regulate quadrupedal robots' locomotion noise in real-world deployments, we aim to learn noise-constrained policy in simulation.
Locomotion noise in quadrupedal robots mainly arises from foot-ground impacts, where mechanical energy is transferred into vibrational and acoustic energy. 
The noise intensity correlates with the impact energy, which is proportional to the squared impact velocity, and the contact force~\cite{akay1978review}. 
Since direct noise level measurements are unavailable in simulation, we design a noise cost function based on contact force and impact velocity.
Specifically, we define the noise cost as:
\begin{equation}
    c(s, a) = \lambda_1 \sum_{i} \exp(\frac{F_{\text{max}}-\|\boldsymbol{F}_i\|}{\sigma_{F}}) + \lambda_2 \sum_{i} \exp(-\frac{v_{\text{impact}, i}^2}{\sigma_v}),
    \label{eq:cost}
\end{equation}
where $\boldsymbol{F}_i$ denotes the contact force for foot $i$ ($i\in \{0, 1, 2, 3\}$), $F_{\text{max}}$ is the predefined maximum force threshold, and $v_{\text{impact}, i}$ represents the impact velocity.
We use scaling factors $\sigma_v=0.5, \sigma_{F}=50$ and $\lambda_1=\lambda_2=0.5$.
Additionally, an exponential penalty function is used to capture the nonlinear scaling of perceived noise in decibels, ensuring strong penalties for large contact forces and impact velocities. 
We visualize the relationship between physics-informed cost and impact forces and impact velocities in Section \ref{sec:exp_setup} Figure \ref{fig:cost_impact_correlation}.

We formulate this noise minimization problem as a Constrained Markov Decision Process (CMDP), defined by the tuple $\mathcal{M} = (\mathcal{S}, \mathcal{A}, P, r, c, \mu_0)$, where $\mathcal{S}$ is the state space, $\mathcal{A}$ is the action space, $P$ is the transition function, $r$ is the reward function, $c$ is the noise cost function, and $\mu_0$ is the initial state distribution. 
The objective of Safe RL is to find an optimal policy $\pi$ that maximizes the expected return while ensuring that the expected cumulative cost remains within a predefined constraint threshold $\epsilon$:


\begin{equation}
    \pi^* = \arg\max_{\pi} J_r^{\pi}(s_0), \quad \text{s.t.} \quad J_c^{\pi}(s_0) \leq \epsilon,
    \label{eq:safe_rl_objective}
\end{equation}

where the expected reward return and cost return are defined as:
\begin{equation}
    J_f^{\pi}(s_0) = \mathbb{E}_{\tau \sim \pi, s_0 \sim \mu_0} \left[\sum_{t=0}^{\infty} \gamma^t f_t\right], \quad f \in \{r, c\}.
    \label{eq:value_function}
\end{equation}

This formulation ensures that the learned policy $\pi^*$ optimally balances locomotion performance while adhering to noise constraints across different deployment conditions.

\subsection{Conditional Policy for Adjustable Noise Reduction}
\label{method:conditional_policy}

Standard Safe RL learns policies optimized for a single pre-defined noise threshold $\epsilon$. However, real-world deployments require dynamic noise adaptation across different environments and user preferences. Training separate policies for each constraint level is computationally inefficient and impractical. 

To address this, we introduce a versatile conditional policy $\pi(\cdot \mid s, \epsilon)$ that dynamically adapts its behavior to different noise constraints $\epsilon$ without retraining. 
Instead of training separate policies for each constraint level, we extend the Safe RL formulation to a constraint-conditioned RL setting, where the policy is conditioned on a distribution of noise thresholds $\epsilon \in \mathcal{E}$. The optimal conditional policy is then defined as:
\begin{equation}
    \pi^*(\cdot \mid s, \epsilon) = \arg\max_{\pi} J_r^{\pi}(s_0), \; 
    \text{s.t.} \; J_c^{\pi}(s_0) \leq \epsilon, \; \forall \epsilon \in \mathcal{E},
    \label{eq:conditioned_policy}
\end{equation}
where $J_r^{\pi}(s_0)$ and $J_c^{\pi}(s_0)$ are the expected cumulative reward and cost, respectively, as defined in Equation~\eqref{eq:value_function}. 
This formulation enables a single policy to generalize across different noise constraints. 
However, directly conditioning policies on $\epsilon$ poses data efficiency and generalization challenges. Simply concatenating $\epsilon$ to observations often leads to poor performance, as the policy struggles to generalize across the entire constraint space. To address this, we develop a structured Successor Feature-based representation, aiming to enhance both generalization and efficiency.

\subsection{Adjustable Noise-reduction Embedding Decomposition}
\label{method:successor_feature}
In Proximal Policy Optimization (PPO)~\cite{schulman2017proximal}, policy (actor network) updates rely on an estimated advantage function~\cite{schulman2015high}, which is estimated using a critic network to reduce variance in policy gradients. The critic network estimates the expected return $J_r^{\pi}(s)$ with the value function: $V(s) = \mathbb{E} \left[\sum_{t=0}^{\infty} \gamma^t r_t \mid s_0 = s \right].$
In Safe RL, a separate cost critic network is introduced to estimate the expected cost return $J_c^{\pi}(s)$ with cost value function :
$V_c(s) = \mathbb{E}_{\pi}  \left[\sum_{t=0}^{\infty} \gamma^t c_t \mid s_0 = s \right].$
In our constraint-conditioned Safe RL, the policy generalizes across a range of noise constraints $\epsilon$. 
Thus, the critic estimation must be conditioned on $\epsilon$, leading to condition-dependent reward and cost critic networks, respectively parameterized by $\phi$ and $\psi$:
\begin{equation}
    V_f^{\theta_f}(s, \epsilon) = \mathbb{E} \left[\sum_{t=0}^{\infty} \gamma^t f_t \mid s, \epsilon \right], \;f \in \{r, c\}, \; \theta_f \in \{\phi, \psi\}.
\end{equation}

To improve generalization across noise constraints, we decompose the condition-aware critic networks using Successor Features (SF), following the convention of~\cite{yao2023constraint}. Instead of directly learning $V_r^\phi(s, \epsilon)$ and $V_c^\psi(s, \epsilon)$, we factorize them into a state-dependent representation $\boldsymbol{\xi}_f^\pi(s)$ and constraint-dependent weight functions $\boldsymbol{w}_f(\epsilon)$, $f \in \{r, c\}$:
\begin{equation} 
V_f^{\theta_f}(s, \epsilon) = \boldsymbol{\xi}_f(s)^\top\boldsymbol{w}_f(\epsilon). \end{equation}

Here, $\boldsymbol{\xi}_f(s)$ encodes a shared representation of state dynamics, independent of the constraint level, while $\boldsymbol{w}_f(\epsilon)$ modulates this representation to adapt to different noise thresholds. This formulation enables a single state representation to generalize across different constraint levels by simply adjusting the respective weight function.

The reward critic and cost critic are optimized to approximate the reward return and cost return:
\begin{equation}
    \mathcal{L}_\text{critic}(\theta_f) = \mathbb{E} \left[\left( V_f^{\theta_f}(s, \epsilon) - J_f^\pi(s, \epsilon) \right)^2 \right].
    \label{eq:critic_loss}
\end{equation}

\subsection{Policy Training}
Our policy is optimized across a range of constraint levels, $\epsilon \in \mathcal{E}$, where $\mathcal{E} = \{\epsilon_i \mid i = 1, \dots, N\}$. Here, $N$ denotes the total number of environments, and each constraint level $\epsilon_i$ is uniformly interpolated over $[0,1]$. Each environment $i$ is assigned a constraint level $\epsilon_i$, with $|\mathcal{E}| = N$.


For policy optimization, we utilize PPO-Lagrangian~\cite{stooke2020responsive}, which introduces a Lagrange multiplier $\lambda$ to dynamically regulate cost constraints. This is achieved by incorporating cost advantage estimation into the overall advantage computation $\hat{A}_t$, as defined in Equation~\ref{eq:lagrangian advantage}.
Unlike standard Safe RL with a single cost threshold, our approach maintains $N$ distinct Lagrangian multipliers $\Lambda = \{\lambda^i \mid i = 1, \dots, N\}$,  $\lambda^i$ for each environment $i$.
They adjust the constraint enforcement for its corresponding policy conditioned on $\epsilon_i$. 
Each rollout policy $\pi^\theta(\cdot \mid s, \epsilon_i)$ conditioned on different $\lambda^i$ maintains separate rollout buffer for Lagrange multiplier $\lambda^i$ updates, advantage $\hat{A}_t^i$ estimation, and surrogate loss $\mathcal{L}_{\text{actor}}^i$ calculations:
\begin{equation}
    \hat{A}_t^i = \frac{1}{1 + \lambda^i} \left[ A_r^i(s_t, a_t, \epsilon^i) - \lambda^i A_c^i(s_t, a_t, \epsilon^i) \right],
    \label{eq:lagrangian advantage}
\end{equation}
where $A_r^i(s_t, a_t, \epsilon^i)$ and $A_c^i(s_t, a_t, \epsilon^i)$ represent the estimated reward advantage and cost advantage for $\pi^\theta(\cdot \mid s, \epsilon_i)$.
The surrogate loss for actor network is averaged across all $ \epsilon_i \in \mathcal{E}$,
\begin{equation}
\mathcal{L}_{\text{actor}}(\theta) = \frac{1}{N} \sum_{i=1}^{N} \mathcal{L}^i(\theta),
\end{equation}
\begin{equation}
\label{equ: L_i}
\mathcal{L}^i(\theta) = \mathbb{E} \left[ \text{min}( \eta_t(\theta)\hat{A}_t^i, \text{clip}(\eta_t(\theta), 1 - \epsilon_{\text{clip}}, 1 + \epsilon_{\text{clip}}) \hat{A}_t^i)\right],\end{equation}
where $\displaystyle{\eta_t(\theta) = \frac{\pi_{\theta}(a_t|s_t, \epsilon_i)}{\pi_{\theta_{\rm old}}(a_t|s_t, \epsilon_i)}}$, is the probability ratio between the new policy $\pi_{\theta}$, and old policy $\pi_{\theta_{\rm old}}$.
The Lagrange multiplier \(\lambda^i\) is updated using a PID controller~\cite{stooke2020responsive}.
The policy learning process is summarized in Algorithm \ref{alg:noise_constrained_rl}.
 
\begin{algorithm}
\caption{Noise-Constrained Locomotion with Constrained RL}
\label{alg:noise_constrained_rl}
\begin{algorithmic}[1]
\REQUIRE Actor $\pi^{\theta}$, Reward and Cost Critics $V_r^{\phi}$ and $V_c^{\psi}$, Constraints $\epsilon_i \in \mathcal{E}$, Lagrangian multipliers $\lambda^i \in \Lambda $
\FOR{iteration $k$ in range of max iterations}
    \FOR{each environment $i$}
        \STATE Sample trajectories $\{\tau\}_i \sim \pi^{\theta}(\cdot\mid s_0, \epsilon_i)$ 
        \STATE Compute advantage $\hat{A}_t^i$ using \eqref{eq:lagrangian advantage}
        \STATE Update Lagrange multipliers 
        \STATE Compute surrogate loss $\mathcal{L}^i(\theta)$ using \eqref{equ: L_i}
    \ENDFOR
    \STATE Mean surrogate loss: 
    $\mathcal{L}_{\text{actor}}(\theta) = \frac{1}{N} \sum_{i=1}^{N} \mathcal{L}^i(\theta)$
    \STATE Update actor parameter $\theta$ using $\mathcal{L}_{\text{actor}}$
    \STATE Update critics parameters $\phi$ and $\psi$ using \eqref{eq:critic_loss}
\ENDFOR
\RETURN policy parameters $\theta$
\end{algorithmic}
\end{algorithm}

\section{Experiments}

\subsection{Experiment Setup}
\label{sec:exp_setup}
\begin{figure}[t]
    \centering
    \includegraphics[width=1.0\linewidth]{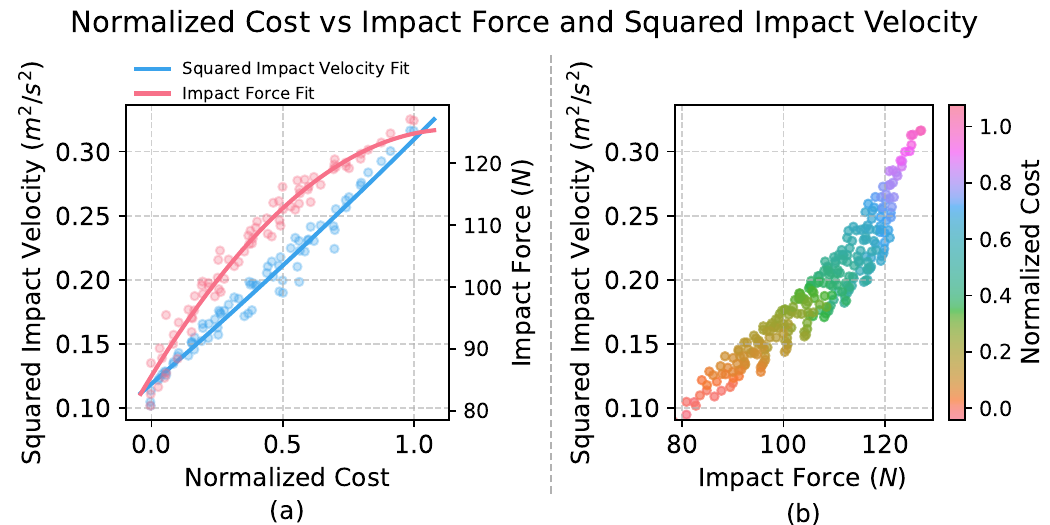}
    \caption{
    (a) Trend lines showing the relationship between Normalized Cost and both Square Impact Velocity $(m^2/s^2)$ and average top $10\%$ Impact Force $(N)$. illustrating a positive correlation. 
    (b) Illustration of the relationship between impact velocity and impact force across different levels of Normalized Cost. As Normalized Cost increases, both squared impact velocity and impact force exhibit corresponding growth.
    }
    \vspace{-10pt}
    \label{fig:cost_impact_correlation}
\end{figure}

\textbf{Simulation setup.}
We use Isaac Gym ~\cite{makoviychuk2021isaac} to train the quiet locomotion policy based on the open-source framework in \cite{margolis2023walk}. 
Reward function are adapted from ~\cite{margolis2023walk}.
The cost function defined in Equation \eqref{eq:cost} is normalized to $[0, 1]$.
We show how normalized cost corresponds to impact velocities and impact force in Figure \ref{fig:cost_impact_correlation}.
Both impact velocity and impact force follow an exponential relationship. However, impact force exhibits stronger non-linearity because its scaled values $\frac{F_{\text{max}}-\|\boldsymbol{F}_i\|}{\sigma_{F}}$ are larger.

The actor-network has hidden dimensions $[512, 256, 128]$.
The critic shares the same hidden dimensions as the actor-network but outputs a $64$-dimensional feature vector $\boldsymbol{\xi}_f(s)$ instead of a scalar value. 
The threshold-conditioned function $\boldsymbol{w}(\epsilon)$ output a $64$-dimensional vector. 
We run 4000 environments in parallel for 10k
iterations on an NVIDIA RTX 4090 GPU, which takes approximately 5 hours.
We adopt a two-stage curriculum for training. We decay the gait-related reward using a sigmoid decay function at the 2k step, first learning a base locomotion policy to track velocity and gait commands, then gradually shifting focus from maintaining a predefined gait to optimizing locomotion for noise minimization.

\textbf{Baselines.} We evaluate our proposed method against several baselines, categorized into conditioned policies and oracle policies. 
Conditioned policies are trained to adapt dynamically to different noise levels by conditioning on a given constraint value: 
\textbf{CONC}: A direct conditioning approach where the noise constraint is concatenated to the policy observation.  
\textbf{R\&C}: A feature expansion method where the noise constraint is repeated 10 times before concatenation, amplifying its representation in the observation space.
Oracle policies are unconditioned and trained separately for each evaluation condition, providing an upper bound on noise reduction performance: \textbf{Oracle SafeRL}~\cite{stooke2020responsive}: A set of 8 PPO-Lagrangian policies, each trained for a specific evaluation condition. These policies serve as an oracle, achieving the best possible balance between cost satisfaction and reward;
\textbf{Oracle Multi-Objective RL (MORL)}~\cite{van2014multi}: A set of 8 PPO policies trained using noise as a reward signal rather than a cost constraint. Each policy employs a different reward scale, following the scalarization approach in~\cite{van2014multi}, to achieve varying levels of noise reduction.

All methods share the same hyperparameters if applicable.

\textbf{Hardware setup.}
We use the Unitree Go2 robot for real-world experiments.  The computations are performed on a host computer. The policy runs at 50Hz and the robot receives the joint position command from the host computer. 
Target joint angles were tracked using a PD controller with gains set to $K_p = 25$ and $K_d = 0.6$. 
Locomotion noise was recorded at a sampling rate of 44.1 kHz. 
To quantify locomotion noise, we compute the Sound Pressure Level (SPL) in decibels (dB SPL) using the Root Mean Square (RMS) amplitude of the recorded audio signal. 
In standard SPL calculations, decibel levels are referenced to a fixed reference pressure:
$P_{\text{ref}} = 2 \times 10^{-5} \text{ Pa},
$
which corresponds to the threshold of human hearing at 1 kHz. 
The SPL is then computed as:
$L_{\text{dB}} = 20 \log{10} \left(\frac{\text{RMS}}{P_{\text{ref}}} \right).$
Since sound intensity is proportional to pressure squared, the logarithmic transformation uses a factor of 20 instead of 10 to correctly represent pressure ratios in decibel units.
To isolate locomotion-induced noise, we first extract the RMS amplitude for both locomotion and environmental noise segments. The environmental noise power is then subtracted before computing the final locomotion noise level.
We evaluate the real-world velocity tracking performance with Prime$^{X}$ 22 cameras, using OptiTrack’s motion capture system.

\subsection{Reward and Cost Violation in Simulation}
\begin{table}[t]
    \centering
    \caption{Comparison of Average Cost Violation and Tracking Error across different terrains.}
    \label{tab:avg_cost_tracking}
    \begin{tabular}{lcc}
        \toprule
        \textbf{Method} & \textbf{Even Terrain} & \textbf{Rough Terrain} \\
        \midrule
        & \multicolumn{2}{c}{\textbf{Avg Normalized Cost Violation} $\downarrow$} \\
        \midrule
        CONC & 0.127 $\pm$ 0.027 & 0.169 $\pm$ 0.038 \\
        R\&C & 0.135 $\pm$ 0.039 & 0.192 $\pm$ 0.029 \\
        Oracles MORL & 0.212 $\pm$ 0.033  &0.265 $\pm$ 0.042\\
        Oracles SafeRL & \textcolor{gray}{\textbf{ 0.013}} $\pm$ 0.001  & \textcolor{gray}{\textbf{0.039}} $\pm$ 0.002 \\
        \textbf{CNCP (ours)} & \textbf{0.107} $\pm$ \textbf{0.018} & \textbf{0.158} $\pm$ \textbf{0.021} \\
        \midrule
        & \multicolumn{2}{c}{\textbf{Avg Tracking Error (m/s)} $\downarrow$} \\
        \midrule
        CONC & 0.348 $\pm$ 0.029 & 0.595 $\pm$ 0.023 \\
        R\&C & 0.334 $\pm$ 0.011 & 0.618 $\pm$ 0.009 \\
        Oracles MORL &  0.385 $\pm$ 0.011 & 0.612 $\pm$ 0.004\\
        Oracles SafeRL & 0.659 $\pm$ 0.076  & 0.874 $\pm$ 0.053\\
        \textbf{CNCP (ours)} & \textbf{0.298} $\pm$ \textbf{0.007} & \textbf{0.575} $\pm$ \textbf{0.014} \\
        \bottomrule
    \end{tabular}
\end{table}
The average normalized cost violation is computed as:
$\mathbb{E}_{{v_\text{target}}, \epsilon} \left[ \max(0, c_i - \epsilon_i) \right]$, 
where $c_i$ represents the incurred cost at evaluation, and $\epsilon_i$ is the corresponding noise constraint.
Our evaluation is conducted across predefined noise constraint levels $\epsilon \in \{0, 0.1, 0.2, 0.3, 0.4, 0.6, 0.8, 1.0\}$. This uneven selection emphasizes stricter noise reduction thresholds, ensuring a focused analysis of constrained locomotion performance.
The average tracking error measures the deviation between the commanded and policy rollout velocities, averaged over all evaluation conditions and target speeds ranging from $0.5$ m/s to $2.25$ m/s.

\begin{figure*}[t]
    \centering
    \includegraphics[width=1\textwidth, page=1, trim = 0cm 15cm 0.2cm 0.cm, clip]{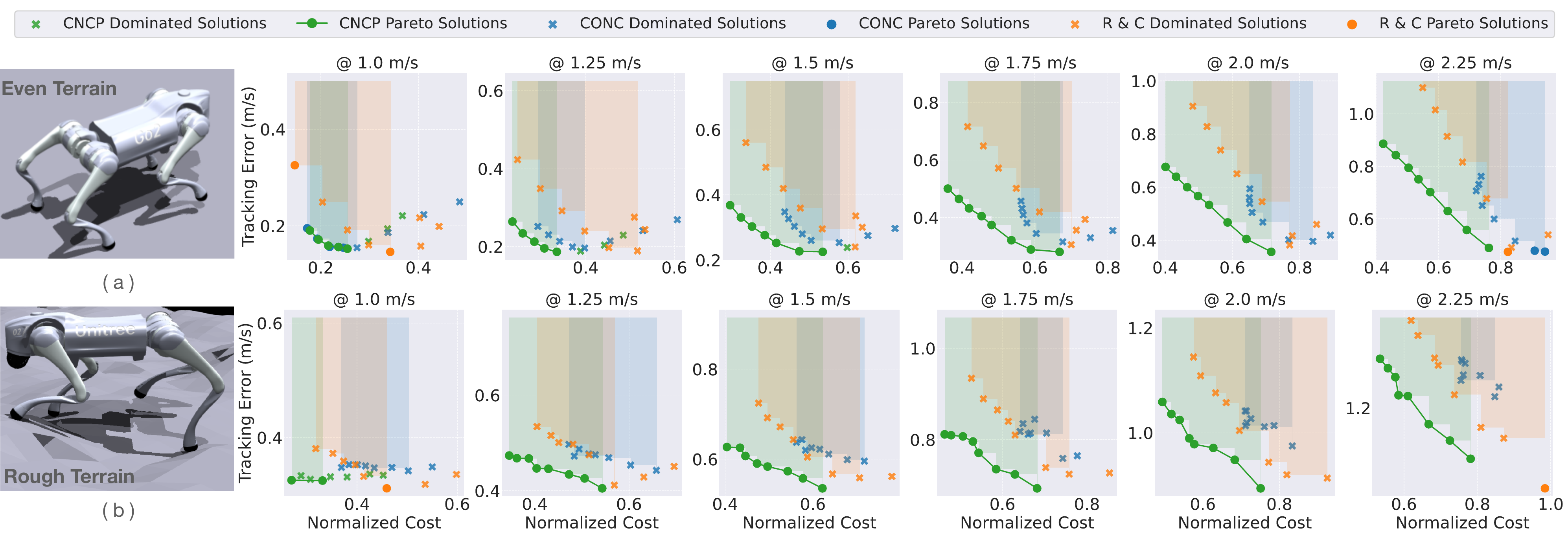}
    \caption{Pareto fronts across target velocities on even (a) and rough (b) terrains. The Pareto front represents solutions that achieve an optimal trade-off between normalized cost and tracking error—where improving one metric degrades the other. Solutions on the front (lower left) are Pareto-optimal, while dominated ones lie above or to the right, indicating suboptimal trade-offs. The shaded area represents the hypervolume~\cite{van2014multi} of each method's Pareto front, with only Pareto solutions contributing to it.}
    \label{fig:pareto_front}
\end{figure*}

As shown in Table~\ref{tab:avg_cost_tracking}, CNCP (ours) achieves the lowest average cost violation among conditioned policies while maintaining the best tracking accuracy. The lowest cost violation and tracking error values are highlighted in \textbf{bold}. Each policy is averaged across three seeds. Compared to CONC and R\&C, CNCP effectively balances noise reduction and velocity tracking, demonstrating the advantage of SF-based conditioning. 
Oracle SafeRL policies, shown in \textcolor{gray}{gray}, are unconditioned and trained separately for each evaluation condition, providing an upper bound on noise reduction performance. While they achieve minimal cost violation, this comes at the cost of significantly higher tracking errors and requiring 8 times more training data.
Oracle MORL does not prioritize noise minimization and lacks explicit adherence to noise constraints.
Overall, CNCP achieves a strong trade-off, dynamically adjusting noise minimization while preserving tracking accuracy.

\label{sec: sim}

\begin{table}[t]
    \centering
    \caption{Comparison of Hypervolume and Sparsity at Different Target Velocities ($\times 10^{-2}$)}
    \label{tab:hv_sparsity}
    \begin{tabular}{c|ccc|ccc}
        \toprule
        \multirow{2}{*}{$V_{x}$(m/s)} & \multicolumn{3}{c|}{Hypervolume $\uparrow$} & \multicolumn{3}{c}{Sparsity $\downarrow$} \\
        & CONC & R\&C & \textbf{CNCP} & CONC & R\&C & \textbf{CNCP} \\
        \midrule
        1.0  & 3.423  & \textbf{5.152}  & 2.586  & \textbf{0.110}  & 1.938  & 0.267  \\
        1.25 & 4.296  & \textbf{8.989}  & 4.048  & \textbf{0.182}  & 0.864  & 0.379  \\
        1.5  & 6.599  & 9.705  & \textbf{11.286}  & 1.804  & \textbf{0.937}  & 1.018  \\
        1.75 & 5.921  & 9.508  & \textbf{15.453}  & 2.048  & 2.309  & \textbf{1.512}  \\
        2.0  & 10.368  & 8.790  & \textbf{14.962}  & 2.839  & 3.429  & \textbf{1.878}  \\
        2.25 & 12.182  & 7.051  & \textbf{14.159}  & 2.410  & 4.446  & \textbf{1.775}  \\
        Avg. & 7.132  & 8.199  & \textbf{10.416}  & 1.566  & 2.321  & \textbf{1.138}  \\
        \bottomrule
    \end{tabular}
\end{table}

\subsection{Pareto Optimality}

\textbf{Pareto Dominance.} A solution is dominant if no other solution outperforms it in all objectives, while a dominated solution is suboptimal~\cite {deb2011multi, van2014multi}. 
In noise-constrained quadrupedal locomotion, dominance reflects the trade-off between stability and velocity tracking. 
The Pareto front captures policies that achieve the best balance, where improving one objective would degrade the other. Figure~\ref{fig:pareto_front} illustrates the Pareto fronts across different target velocities, comparing our method with conditioned baselines.
CNCP dominates a significant portion of CONC solutions and R\&C solutions are dominated by CNCP, highlighting the effectiveness of successor feature decomposition.

Both CONC and R\&C struggle to achieve competitive solutions, leading to suboptimal trade-offs with either poor constraint satisfaction or excessive tracking errors.  
In contrast, CNCP learns a superior \textit{Pareto front}, with the best solutions concentrated in the lower-left corner, minimizing both cost and tracking error. 
By explicitly modeling the interaction between state dynamics and constraints, CNCP dynamically adjusts locomotion behavior across different noise levels. 
This advantage becomes even more significant at higher velocities, where naive conditioning methods fail to generalize.  
By leveraging a decomposed representation, CNCP captures how the constraint-dependent weight function $w(\epsilon)$ modulates the state representation $\psi^\pi(s)$, enabling adaptive noise-aware control and yielding Pareto-optimal solutions in conditional safe RL.

\textbf{Hypervolume and Sparsity.}
To quantitatively assess Pareto front quality, we adopt hypervolume~\cite{van2014multi} and sparsity~\cite{xu2020prediction} as performance measures, following.  
We compute the hypervolume and sparsity for each method using the set of solutions obtained from its policy rollouts across different conditions.  
A higher hypervolume (bolded in Table~\ref{tab:hv_sparsity}) indicates a more dominant Pareto front, reflecting superior trade-offs between locomotion performance and noise minimization. Meanwhile, a lower sparsity (also bolded in Table~\ref{tab:hv_sparsity}) signifies a more uniformly distributed set of Pareto-optimal solutions.
As shown in the table, CNCP consistently achieves the highest hypervolume on average and across most velocities, demonstrating superior cost-performance trade-offs. 
Sparsity quantifies the uniformity of Pareto-optimal solutions by assessing the variation in spacing between consecutive points along the front. 
It is computed as the standard deviation of distances between adjacent solutions in the Pareto set, with lower values indicating a more evenly distributed front. 
CNCP achieves the lowest sparsity on average and in half of the evaluation cases while remaining competitive in the rest, suggesting well-balanced solution spacing.

\begin{figure}[tbp]
    \centering
    \includegraphics[width=0.48\textwidth, page=1, trim = 0.2cm 0cm 0.5cm 0.cm, clip]{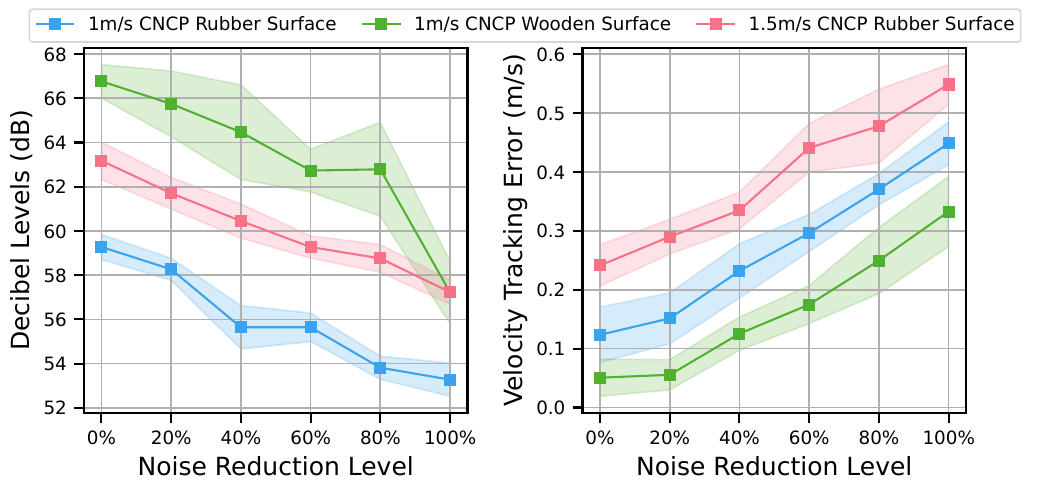}
    \vspace{-15pt}
    \caption{Real-world evaluation of CNCP under different terrains and target velocities. The 1m/s deployment runs for 300 steps (6s at 50Hz), while the 1.5m/s deployment runs for 200 steps (4s).}
    \label{fig:real_terrain}
    \vspace{-10pt}
\end{figure}

\subsection{Real World Adjustable Noise Reduction}
\label{sec: real}
In the real-world deployment, we analyze the Go2 robot locomotion noise and velocity tracking performance to showcase the effectiveness of our conditioned noise-constrained policy. 
During deployment, the $\epsilon$ acts as a noise reduction effect steering command, allowing real-time adjustment of the noise minimization. 
Locomotion sounds are recorded using a smartphone microphone securely mounted on the robot’s back to ensure consistent noise measurement and decibel levels are calculated as described in \ref{sec:exp_setup}. 
Velocity tracking errors are measured using Prime$^{X}$ 22 cameras with OptiTrack’s motion capture system.

\begin{figure}[t]
    \centering
    \includegraphics[width=0.48\textwidth, page=1, trim = 0.2cm 0cm 0.5cm 0.cm, clip]{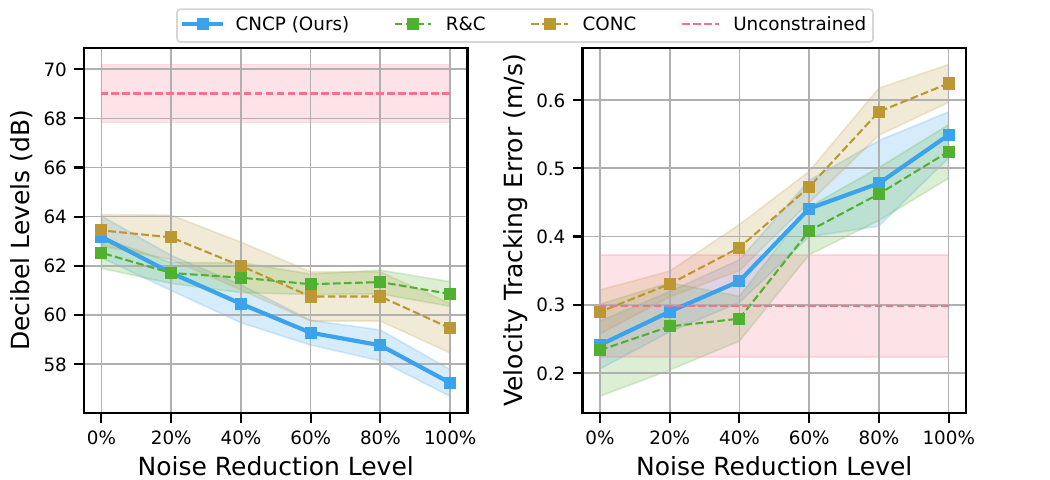}
    \vspace{-15pt}
    \caption{Real-world comparison of CNCP with baseline conditional policies and an unconstrained policy. Each deployment runs for 200 steps (4s) on rubber terrain at 1.5 m/s. The unconstrained policy remains constant across conditions as it does not take noise reduction level as input.}
    \label{fig:real_baseline}
    \vspace{-5pt}
\end{figure}

\begin{figure}[t]
    \centering
    \includegraphics[width=0.48\textwidth, page=1, trim = 0cm 0cm 11.5cm 0.cm, clip]{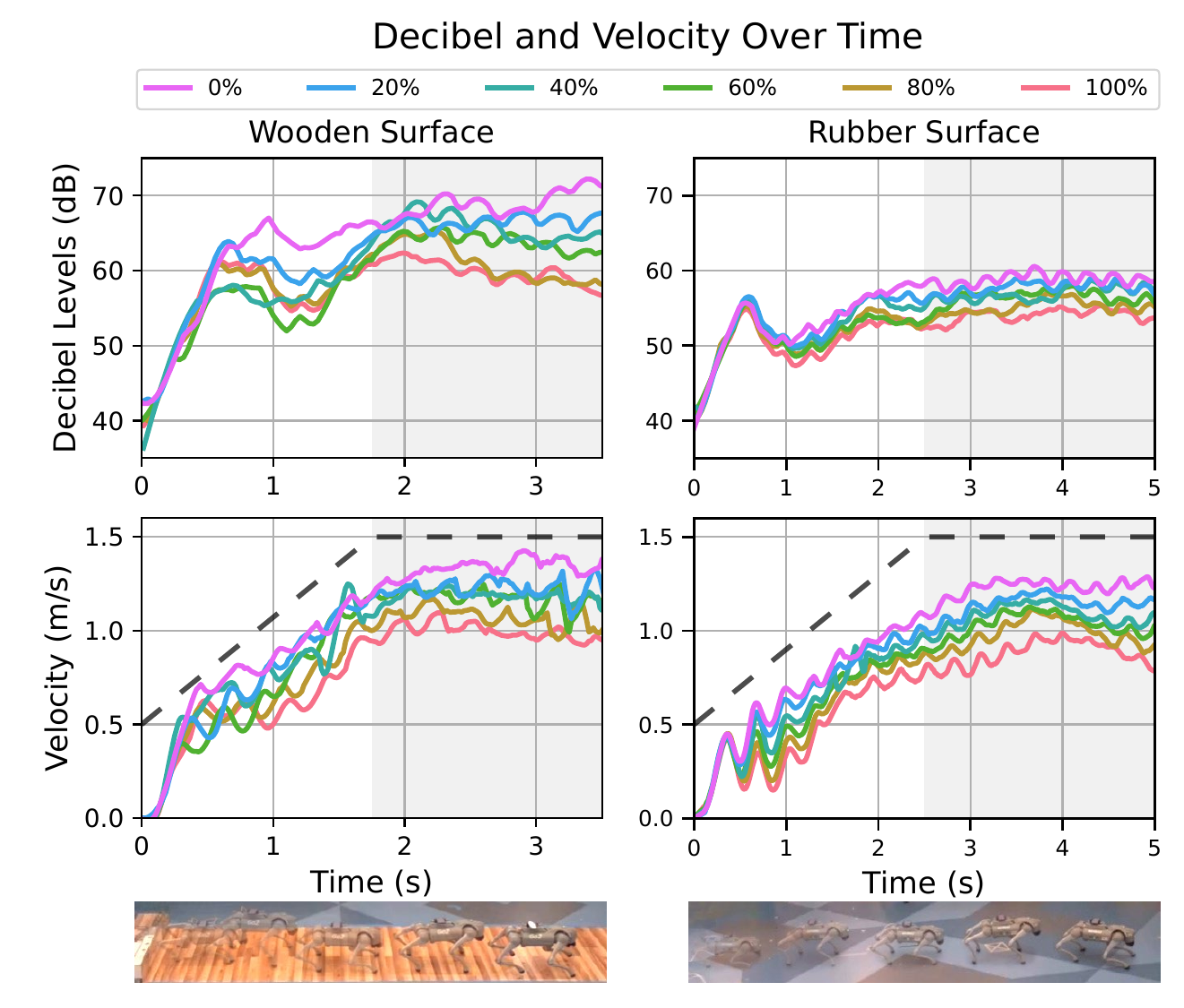}
    \vspace{-20pt}
    \caption{Decibel levels and velocity tracking over time on wooden flooring (left) and rubber flooring (right). The target velocity increases from 0.5 to 1.5 m/s in the first half and maintains 1.5 m/s in the second half.}
    \vspace{-5pt}
    \label{fig:real_wood}
\end{figure}

In Figure~\ref{fig:real_terrain}, we evaluate our conditional policy under different conditions: 1m/s on rubber surface, 2m/s on the rubber surface, and 1m/s on the wooden surface. The results show that our policy effectively reduces noise as the noise reduction level increases, corresponding to the conditioning variable $\epsilon$ in our policy. As expected, higher velocities generate more noise, and the wooden surface produces greater noise levels than the rubber surface. Since noise characteristics vary across terrains and target velocities, a conditional policy is essential for adjustable noise reduction strategies in different deployment settings.   
Figure~\ref{fig:real_baseline} compares our method to baseline conditional policies, demonstrating that CNCP achieves more significant noise reduction and maintains relatively low tracking error.
While all constrained policies exhibit quieter locomotion compared to the unconstrained policy, CNCP achieves a more substantial noise reduction across different conditions.  
Our method also achieves a better balance between noise reduction and tracking error compared to baselines. 
This validates the effectiveness of successor feature-based decomposition for conditional noise reduction, achieving a wider range of noise adjustments while maintaining a better trade-off with velocity tracking.

In Figure~\ref{fig:real_wood}, the robot follows an increasing target velocity on wooden and rubber surfaces, showcasing CNCP's ability to adjust noise levels based on the conditioned noise reduction level.  
Across different target velocities and terrains, higher noise reduction levels consistently lead to lower decibel levels, demonstrating effective noise reduction. The trade-off between noise minimization and velocity tracking is evident, as stricter noise constraints cause slight deviations from the target velocity. In contrast, CNCP maintains stable tracking performance, ensuring efficient locomotion while effectively reducing noise.

\section{Conclusion}

In this work, we introduced the Conditional Noise-Constrained Policy (CNCP) for adaptive noise-aware quadrupedal locomotion. 
Our method enables flexible noise control by conditioning the policy on noise constraints, allowing seamless adaptation to different deployment settings without retraining. 
By leveraging Successor Feature decomposition within the PPO critic network, we disentangle state-dependent representations from constraint-dependent adaptation. 
CNCP demonstrates a superior Pareto trade-off between locomotion agility and noise minimization in simulations while achieving effective noise control in real-world experiments.
Our results highlight the flexibility of conditional policy for real-world adaptable and socially aware quadrupedal deployment.
Future work will explore extending CNCP to more complex environments and integrating human-in-the-loop steering for personalized noise adaptation.


\bibliographystyle{IEEEtran}
\bibliography{root}  

\end{document}